\documentclass[runningheads]{llncs}

\usepackage{times}
\usepackage{url}
\usepackage{helvet}
\usepackage{courier}

\usepackage{graphicx}
\usepackage{amsfonts}
\usepackage{amssymb}
\usepackage{ifthen}
\usepackage{verbatim}
\usepackage{xspace}
\usepackage{pstricks}
\usepackage{pst-tree}

\newcommand{\mxsat}{MaxSAT\xspace}

\newcommand{\DNote}[1]{\ifthenelse{\boolean{draftmode}}{\medskip{\noindent{\bf{Note}:} {\em #1}}\medskip}{}}
\newenvironment{Proof}{{\noindent{\bf Proof:}}\normalfont\slshape}{\hfill\rule{1.5mm}{1.5mm}} 
\newboolean{submode}            
\setboolean{submode}{true}      
\newboolean{draftmode}          
\setboolean{draftmode}{true}    

\urldef{\mailsa}\path|jpms@ecs.soton.ac.uk|
\urldef{\mailsb}\path|{ines,vmm}@sat.inesc-id.pt|

\begin{document}

\title{Symmetry Breaking for Maximum Satisfiability\thanks{This paper
    is also available as reference~\cite{sat-tr039-08}.}}

\author{Joao Marques-Silva\inst{1} \and In\^{e}s Lynce\inst{2} \and
  Vasco Manquinho\inst{2}}
\institute{School of Electronics and Computer Science, University of
  Southampton, UK\\\mailsa\\\and
  IST/INESC-ID, Technical University of Lisbon, Portugal\\\mailsb
}
\titlerunning{Symmetry Breaking for MaxSAT}
\authorrunning{Marques-Silva, Lynce and Manquinho}

\maketitle


\begin{abstract}
Symmetries are intrinsic to many combinatorial problems including
Boolean Satisfiability (SAT) and Constraint Programming (CP).
In SAT, the identification of symmetry breaking predicates~(SBPs) is a
well-known, often effective, technique for solving hard problems.
The identification of SBPs in SAT has been the subject of significant
improvements in recent years, resulting in more compact SBPs and more
effective algorithms.
The identification of SBPs has also been applied to pseudo-Boolean
(PB) constraints, showing that symmetry breaking can also be an
effective technique for PB constraints.
This paper extends further the application of SBPs, and shows that
SBPs can be identified and used in Maximum Satisfiability (\mxsat),
as well as in its most well-known variants, including partial \mxsat, 
weighted \mxsat and weighted partial \mxsat.
As with SAT and PB, symmetry breaking predicates for \mxsat and
variants are shown to be effective for a representative number of
problem domains, allowing solving problem instances that current state
of the art \mxsat solvers could not otherwise solve.
\end{abstract}



\section{Introduction}
\label{sec:intro}

Symmetry  breaking is a widely used technique for
solving combinatorial problems.
Symmetries have been used with great success in
Satisfiability~(SAT)~\cite{crawford-kr96,kas-ijcai03}, and 
are regarded as an essential technique for solving specific
classes of problem instances.
Symmetries have also been widely used for solving constraint satisfaction problems~(CSPs)~\cite{sym-handbook06}.
More recent work has shown how to apply symmetry 
breaking in pseudo-Boolean~(PB)
constraints~\cite{kas-aspdac04} and also in soft 
constraints~\cite{smith-cp07}.
It should be noted that symmetry  breaking is viewed
as an effective problem solving technique, either for SAT, PB or CP,
that is often used as an alternative technique, to be applied when
default algorithms are unable to solve a given problem instance.

In recent years there has been a growing interest in algorithms for
\mxsat and
variants~\cite{LMP05,LMP06,zhang-aij05,heras-aaai06,HLO07,su-ijcai07},
in part because of the wide range of potential applications. 
\mxsat and variants represent a more general framework than either SAT
or PB, and so can naturally be used in many practical applications.
The interest in \mxsat and variants motivated the development of a new
generation of \mxsat algorithms, remarkably more efficient than early
\mxsat algorithms~\cite{freuder-dimacs96,borchers-jco98}. Despite the
observed improvements, there are many problems still too complex 
for \mxsat algorithms to solve~\cite{maxsatcomp}. 
Natural lines of research for improving \mxsat algorithms include
studying techniques known to be effective for either SAT, PB or
CP. One concrete example is symmetry  breaking.
Despite its success in SAT, PB and CP, the usefulness of symmetry
breaking for \mxsat and variants has not been thoroughly studied.
%

This paper addresses the problem of using symmetry breaking in \mxsat
and in its most well-known variants, partial \mxsat, weighted \mxsat
and weighted partial \mxsat. The work extends past recent work on
computing symmetries for SAT~\cite{kas-ijcai03} and PB 
constraints~\cite{kas-aspdac04} by computing automorphism on 
colored graphs obtained from CNF or PB formulas, 
and by showing how symmetry breaking
predicates~\cite{crawford-kr96,kas-ijcai03} can be exploited.
The experimental results show that symmetry breaking is an effective
technique for \mxsat and variants, allowing solving problem instances
that state of the art \mxsat solvers could not otherwise solve.

The paper is organized as follows. 
The next section introduces the notation used throughout the paper,
provides a brief overview of \mxsat and variants, and also summarizes
the work on symmetry breaking for SAT and PB constraints. Afterwards,
the paper describes how to apply symmetry  breaking in \mxsat and
variants. Experimental results, obtained on representative problem
instances from the \mxsat evaluation~\cite{maxsatcomp} and also from
practical applications~\cite{kas-ijcai03}, demonstrate that symmetry
breaking allows solving problem instances that could not be solved by
{\em any} of the available state of the art \mxsat solvers. The paper
concludes by summarizing related work, by overviewing the main
contributions, and by outlining directions for future work.


\section{Preliminaries}
\label{sec:prelim}


This section introduces the notation used through the paper,
as well as the \mxsat problem and its variants. An overview
of symmetry identification and symmetry breaking is also 
presented.

\subsection{Maximum Satisfiability}
\label{ssec:mxsat}


The paper assumes the usual definitions for SAT. A propositional
formula is represented in {\em Conjunctive Normal Form}~(CNF). A CNF
formula $\varphi$
consists of a conjunction of clauses, where each clause $\omega$ is a
disjunction of literals, and a literal $l$ is either a propositional
variable $x$ or its complement $\bar{x}$.
Variables can be assigned a propositional value, either 0 or 1.
A literal $l_j = x_j$ assumes value 1 if $x_j = 1$ and assumes value 0
if $x_j = 0$. Conversely, literal $l_j = \bar{x}_j$ assumes value 1 if
$x_j = 0$ and value 0 when $x_j = 1$.
For each assignment of values to the variables, the value of formula
$\varphi$ is computed with the rules of propositional logic.
A clause is said to be {\em satisfied} if at least one of its literals
assumes value 1. If all literals of a clause assume value 0, then the
clause is {\em unsatisfied}.
The propositional satisfiability (SAT) problem consists in deciding
whether there exists an assignment to the variables such that $\varphi$
is satisfied.


Given a propositional formula $\varphi$, the \mxsat problem is defined
as finding an assignment to variables in $\varphi$ such that the 
number of satisfied clauses is maximized. (\mxsat can also be defined 
as finding an assignment that minimizes the number of unsatisfied clauses.)
Well-known variants of \mxsat include partial \mxsat, weighted \mxsat
and weighted partial \mxsat. 
%

For partial \mxsat, a propositional formula $\varphi$ is described by
the conjunction of two CNF formulas $\varphi_s$ and $\varphi_h$, where
$\varphi_s$ represents the {\em soft} clauses and $\varphi_h$
represents the {\em hard} clauses.
The partial \mxsat problem over a propositional formula $\varphi =
\varphi_h \wedge \varphi_s$ consists in finding an assignment to the
problem variables such that all hard clauses ($\varphi_h$) are
satisfied and the number of satisfied soft clauses ($\varphi_s$) is
maximized.

For {\em weighted} \mxsat, each clause in the CNF formula is
associated to a non-negative weight. A weighted clause is a 
pair $(\omega, c)$ where $\omega$ is a classical clause and 
$c$ is a natural number corresponding to the cost of unsatisfying
$\omega$. Given a weighted CNF formula $\varphi$, the {\em weighted}
\mxsat problem consists in finding an assignment to problem variables
such that the total weight of the unsatisfied clauses is minimized,
which implies that the total weight of the satisfied clauses is maximized. 
For the {\em weighted partial} \mxsat problem, the formula is the 
conjunction of a weighted CNF formula (soft clauses) and a 
classical CNF formula (hard clauses). The weighted partial  \mxsat
problem consists in finding an assignment to the variables such 
that all hard clauses are satisfied and the total weight of 
satisfied soft clauses is maximized. Observe that, for both partial
\mxsat and weighted partial \mxsat, hard clauses can be represented as
weighted clauses. For these clauses one can consider that the weight is
greater than the sum of the weights of the soft clauses.

\mxsat and variants find a wide range of practical applications,
that include scheduling, routing, bioinformatics, and design
automation. Moreover, \mxsat can be used for solving pseudo-Boolean
optimization~\cite{HLO07}.
The practical applications of \mxsat motivated recent interest in
developing more efficient algorithms. The most efficient algorithms
for \mxsat and variants are based on branch and bound search, using
dedicated bounding and inference
techniques~\cite{LMP05,LMP06,heras-aaai06,HLO07}. 
Lower bounding techniques include for example the use of unit
propagation for identifying necessarily unsatisfied clauses, whereas
inference techniques can be viewed as restricted forms of resolution,
with the objective of simplifying the problem instance to solve.
%

\subsection{Symmetry Breaking}
\label{ssec:symsat}


Given a problem instance, a symmetry is an operation that preserves
the constraints, and therefore also preserves the
solutions~\cite{cohen-cp05}. For a set of symmetric states, it is
possible to obtain the whole set of states from any of the
states. Hence, symmetry breaking predicates may eliminate all but one
of the equivalent states. Symmetry breaking is expected to speed up
the search as the search space gets reduced. For specific problems
where symmetries may be easily found this reduction may be
significant. Nonetheless, the elimination of symmetries necessarily
introduces overhead that is expected to be negligible when compared
with the benefits it may provide.

The elimination of symmetries has been extensively studied in CP and
SAT~\cite{puget-ismis93,crawford-kr96}.
The most well-know strategy for eliminating symmetries in SAT consists
in adding symmetry breaking predicates (SBPs) to the CNF
formula~\cite{crawford-kr96}. 
SBPs are added to the formula before the search starts. The symmetries
may be identified for each specific problem, and in that case it is
required that the symmetries in the problem are identified when creating
the encoding. Alternatively, one may give a formula to a specialized
tool for detecting all the symmetries~\cite{kas-ijcai03}. The
resulting SBPs are intended to merge symmetries in equivalent classes.
In case all symmetries are broken, only one assignment, instead of $n$ 
assignments, may satisfy a set of constraints, being $n$ the
number of elements in a given equivalent class.

Other approaches include remodeling the problem~\cite{smith-leeds01}
and breaking symmetries during search~\cite{gent-ecai00}. Remodeling
the problem implies creating a different encoding, e.g. obtained by
defining a different set of variables, in order to create a problem
with less symmetries or even none at all. Alternatively, the search
procedure may be adapted for adding SBPs as the search proceeds to
ensure that any assignment symmetric to one assignment already
considered will not be explored in the future, or by performing checks
that symmetric equivalent assignments have not yet been visited.

Currently available tools for detecting and breaking symmetries for
a given formula are based on group theory. From each formula a group
is extracted, where a group is a set of permutations. A permutation is
a one-to-one correspondence between a set and itself. Each symmetry
defines a permutation on a set of literals. In practice, each
permutation is represented by a product of disjoint cycles. Each cycle
$(l_1 \: l_2 \: \ldots \: l_m)$ with size $m$ stands for the
permutation that maps $l_i$ on $l_{i+1}$ (with $1 \leq i \leq m-1$)
and $l_m$ on $l_1$. Applying a permutation to a formula will produce
exactly the same formula.

\begin{example}
\label{ex:cnf1}
Consider the following CNF formula:
\[\varphi = (x_1\vee x_2)\wedge(\bar{x}_1\vee
x_2)\wedge(\bar{x}_2)\wedge(x_3\vee x_2)\wedge(\bar{x}_3\vee x_2)\]
The permutations identified for $\varphi$ are $(x_3\:\bar x_3)$ and
$(x_1\:x_3)(\bar x_1\:\bar x_3)$. (The permutation $(x_1\:\bar x_1)$
is implicit.) The formula resulting from the permutation $(x_3\:\bar
x_3)$ is obtained by replacing every occurrence of $x_3$ by $\bar x_3$
and every occurrence of $\bar x_3$ by $x_3$. Clearly, the obtained
formula is equal to the original formula. The same happens when
applying the permutation $(x_1\:x_3)(\bar x_1\: \bar x_3)$: replacing
$x_1$ by $x_3$, $x_3$ by $x_1$, $\bar x_1$ by $\bar x_3$ and $\bar
x_3$ by $\bar x_1$ produces the same formula.
\end{example}


\section{Symmetry Breaking for MaxSAT}
\label{sec:symxsat}

This section describes how to apply symmetry breaking in
\mxsat. First, the construction process for the graph representing 
a CNF formula is briefly reviewed~\cite{crawford-kr96,kas-ijcai03}, 
as it will be modified later in this section. Afterwards, plain 
\mxsat is considered. The next step is to address partial, 
weighted and weighted partial \mxsat.

\subsection{From CNF Formulas to Colored Graphs}

Symmetry breaking for \mxsat and variants requires a few modifications
to the approach used for
SAT~\cite{crawford-kr96,kas-ijcai03}. This section
summarizes the basic approach, which is then extended in the following
sections. 

Given a graph, the {\em graph automorphism} problem consists in
finding isomorphic groups of edges and vertices with a one-to-one 
correspondence. In case of graphs with colored vertices, the
correspondence is made between vertices with the same color.
It is well-known that symmetries in SAT can be identified by 
reduction to a graph automorphism
problem~\cite{crawford-kr96,kas-ijcai03}. 
The propositional formula is represented as an undirected graph
with colored vertices, such that the automorphism in the graph
corresponds to a symmetry in the propositional formula.

Given a propositional formula $\varphi$, a colored undirected graph is
created as follows:
\begin{itemize}
\item For each variable $x_j \in \varphi$ add two vertices to represent 
  $x_j$ and $\bar{x}_j$. All vertices associated with variables are
  colored with color 1;
\item For each variable $x_j \in \varphi$ add an edge between the
  vertices representing $x_j$ and $\bar{x}_j$;
\item For each binary clause $\omega_i = (l_j \vee l_k) \in \varphi$, 
  add an edge between the vertices representing $l_j$ and $l_k$;
\item For each non-binary clause $\omega_i \in \varphi$ create a vertex colored
  with 2;
\item For each literal $l_j$ in a non-binary clause $\omega_i$, add an edge
  between the corresponding vertices.
\end{itemize}

\begin{figure}[t]
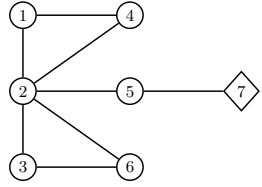

  \begin{center}
  \scalebox{0.7}{
$
\psmatrix[colsep=1.5cm,rowsep=.75cm,mnode=circle] 
 1  & 4  \\
 2  & 5 & [mnode=dia] 7  \\ 
 3  & 6
\psset{arrows=-,arrowscale=2,arrowinset=0}
\ncline{1,1}{1,2}
\ncline{2,1}{2,2}
\ncline{3,1}{3,2}
\ncline{1,1}{2,1}
\ncline{2,1}{3,1}
\ncline{2,1}{1,2}
\ncline{2,1}{3,2}
\ncline{2,2}{2,3}
\endpsmatrix 
$
}
  \end{center}
  \caption{Colored graph for Example~\ref{ex:cgraph1}}
  \label{fig:graph1}
\end{figure}

\begin{example}
\label{ex:cgraph1}
Figure~\ref{fig:graph1} shows the colored undirected graph associated
with the CNF formula of Example~\ref{ex:cnf1}. Vertices with shape $\circ$ represent 
color 1 and vertices with shape $\diamond$ represent color 2. Vertex
1 corresponds to $x_1$, 2 to $x_2$, 3 to $x_3$, 4 to $\bar x_1$, 5 to $\bar x_2$,
6 to $\bar x_3$ and 7 to unit clause $(\bar x_2)$. Edges 1-2, 2-3, 2-4 and 2-6 
represent binary clauses and edges 1-4, 2-5 and 3-6 link complemented literals.
\end{example}

\subsection{Plain Maximum Satisfiability}
\label{ssec:symxsat}



Let $\varphi$ represent the CNF formula of a \mxsat instance.
Moreover, let $\varphi_{sbp}$ be the CNF formula for the
symmetry-breaking predicates obtained with a CNF symmetry tool
(e.g.~Shatter~\footnote{http://www.eecs.umich.edu/$\sim$faloul/Tools/shatter/}). 
All clauses in $\varphi$ are effectively {\em soft}
clauses, for which the objective is to maximize the number of satisfied
clauses. In contrast, the clauses in $\varphi_{sbp}$ are {\em
  hard} clauses, which must necessarily be satisfied. As a result, the
original \mxsat problem is transformed into a partial \mxsat problem,
where $\varphi$ denotes the soft clauses and $\varphi_{sbp}$
denotes the hard clauses. The solution of the partial \mxsat problem
corresponds to the solution of the original \mxsat problem.

\begin{example}
\label{ex:sbp1}
For the CNF formula of Example~\ref{ex:cnf1}, the generated SBP
predicates (by Shatter) are:
$\varphi_{sbp} = (\bar{x}_3)\wedge(\bar{x}_1\vee x_3)$
As result, the resulting instance of partial \mxsat will be
$\varphi' = (\varphi_h\wedge\varphi_s) = (\varphi_{sbp}\wedge\varphi)$.
Moreover, $x_3=0$ and $x_1=0$ are necessary assignments, and so
variables $x_1$ and $x_3$ can be ignored for maximizing the number of
satisfied soft clauses.
\end{example}

Observe that the hard clauses represented by $\varphi_{sbp}$ do
not change the solution of the original \mxsat problem. Indeed, the
construction of the symmetry breaking predicates guarantees that the
maximum number of satisfied soft clauses remains unchanged by the
addition of the hard clauses.

\begin{proposition}
The maximum number of satisfied clauses for the \mxsat problem
$\varphi$ and the partial \mxsat problem $(\varphi\wedge
\varphi_{sbp})$ are the same.
\end{proposition}

\begin{Proof} (Sketch)
The proof follows from the fact that symmetries map models into
models and non-models into non-models (see Proposition~2.1
in~\cite{crawford-kr96}).
Consider the clauses as an ordered sequence $\langle
\omega_1,\ldots,\omega_m\rangle$. Given a symmetry, a clause
in position $i$ will be mapped to a clause in another position
$j$. Now, given any assignment, if the clause in position $i$ is
satisfied (unsatisfied), then by applying the symmetry, the clause in
position $j$ is now satisfied (unsatisfied). Thus the number of
satisfied (unsatisfied) clauses is unchanged.
\end{Proof}

\subsection{Partial and Weighted Maximum Satisfiability}
\label{ssec:symxpsat}


For partial \mxsat, the generation of SBPs needs to be modified. The
graph representation of the CNF formula must take into account the
existence of hard and soft clauses, which must be distinguished by a
graph automorphism algorithm. Symmetric states for problem instances
with hard and soft clauses establish a correspondence either between
hard clauses or between soft clauses. In other words, when applying a
permutation hard clauses can only be replaced by other hard clauses,
and soft clauses by other soft clauses.
In order to address this issue, the colored graph generation needs to
be modified. In contrast to the \mxsat case, binary clauses are not
distinguished from other clauses, and are represented as vertices in
the colored graph.
Clauses can now have one of two colors. A vertex with color 2 is
associated with each soft clause, and a vertex with color 3 is
associated with each hard clause. This modification ensures that any
identified automorphism guarantees that soft clauses correspond only
to soft clauses, and hard clauses correspond only to hard clauses.
Moreover, the procedure for the generation of SBPs from the groups
found by a graph automorphism tool remains unchanged, and the SBPs can
be added to the original instance as {\em new} hard clauses. The
resulting instance is also an instance of partial \mxsat.
Correctness of this approach follows form the correctness of the plain
\mxsat case.

%
The solution for weighted \mxsat and for weighted partial \mxsat is
similar to the partial \mxsat case, but now clauses with different
weights are represented by vertices with different colors.
This guarantees that the groups found by the graph automorphism tool
take into consideration the weight of each clause.
%
Let $\{c_1, c_2, \ldots, c_k\}$ denote the distinct clause weights in
the CNF formula. Each clause of weight $c_i$ is associated with a
vertex of color $i+1$ in the colored graph. In case there exist hard
clauses, an additional color $k+2$ is used, and so each hard clause is
represented by a vertex with color $k+2$ in the colored graph. 
Associating distinct clause weights with distinct colors guarantees
that the graph automorphism algorithm can only make the correspondence 
between clauses with the same weight.
Moreover, the identified SBPs result in new {\em hard} clauses that
are added to the original problem. For either weighted \mxsat or
weighted partial \mxsat, the result is an instance of weighted partial
\mxsat.
As before, correctness of this approach follows form the correctness
of the plain \mxsat case.

\begin{figure}[t]
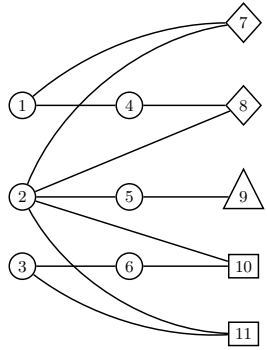

  \begin{center}
  \scalebox{0.7}{
$
\psmatrix[colsep=1.5cm,mnode=r,rowsep=.75cm,mnode=circle] 
    &      &  [mnode=dia] 7  \\
 1  &   4  &  [mnode=dia] 8  \\
 2  &   5  &  [mnode=tri] 9  \\ 
 3  &   6  &  [mnode=r,name=A]\psframebox{10}  \\ 
    &      &  [mnode=r,name=B]\psframebox{11}
\psset{arrows=-,arrowscale=2,arrowinset=0}
\ncline{2,1}{2,2}
\ncline{3,1}{3,2}
\ncline{4,1}{4,2}
\ncline{3,2}{3,3}
\ncarc[arcangle=20]{2,1}{1,3}
\ncline{2,2}{2,3}
\ncarc[arcangle=-20]{4,1}{B}
\ncline{4,2}{A}
\ncarc[arcangle=30]{3,1}{1,3}
\ncline{3,1}{2,3}
\ncline{3,1}{4,3}
\ncarc[arcangle=-30]{3,1}{5,3}
\endpsmatrix 
$
}
  \end{center}
  \caption{Colored graph for Example~\ref{ex:cgraph2}}
  \label{fig:graph2}
\end{figure}

\begin{example}
\label{ex:cgraph2}
Consider the following weighted partial \mxsat instance:
\begin{eqnarray*}
\varphi &=& (x_1\vee x_2, 1)\wedge(\bar{x}_1\vee x_2, 1)\wedge(\bar{x}_2, 5)\wedge \\
&& (\bar{x}_3\vee x_2,9)\wedge(x_3\vee x_2,9)
\end{eqnarray*}
for which the last two clauses are hard. Figure~\ref{fig:graph2} shows
the colored undirected graph associated with the formula. Clauses with
different weights are represented with different colors (shown in the
figure with different vertex shapes). A graph automorphism algorithm
can then be used to generate the symmetry breaking predicates
$\varphi_{sbp}=(\bar{x}_1)\wedge(\bar{x}_3)$, consisting of two hard
clauses. As a result, the assignments $x_1=0$ and $x_3=0$ become
necessary. 
\end{example}

Table~\ref{tab:mods} summarizes the problem transformations described
in this section, where MS represents plain \mxsat, PMS represents
partial \mxsat, WMS represents weighted \mxsat, and WPMS represents
weighted partial \mxsat. 
\begin{table}[t]
\begin{center}
\caption{Problem transformations due to SBPs}
\label{tab:mods}
\begin{tabular}{|c|c|} \hline
Original & With Symmetries \\ \hline\hline
MS       & PMS \\
PMS      & PMS \\ \hline
WMS      & WPMS \\
WPMS     & WPMS \\ \hline
\end{tabular}
\end{center}
\end{table}
The use of SBPs introduces a number of hard clauses, and so the
resulting problems are either partial \mxsat or weighted partial
\mxsat.


\section{Experimental Results}
\label{sec:res}


The experimental setup has been organized as follows. First, all the
instances from the first and second \mxsat evaluations (2006 and
2007)~\cite{maxsatcomp} were run.
These results allowed selecting relevant benchmark families, for which
symmetries occur and which require a non-negligible amount of time for
being solved by both approaches (with or without SBPs). Afterwards,
the instances for which both approaches aborted were removed from the
tables of results. This resulted in selecting the {\tt hamming} and the
{\tt MANN} instances for plain \mxsat, the {\tt ii32} and again the
{\tt MANN} instances for partial \mxsat, the {\tt c-fat500} instances
for weighted \mxsat and the {\tt dir} and {\tt log} instances for
weighted partial \mxsat.

Besides the instances that participated in the \mxsat competition, we
have included additional problem instances ({\tt hole}, {\tt Urq} and
{\tt chnl}). The {\tt hole} instances refer to the well-known pigeon
hole problem, the {\tt Urq} instances represent randomized instances
based on expander graphs and the {\tt chnl} instances model the
routing of wires in the channels of field-programmable integrated
circuits. These instances refer to problems that can be naturally
encoded as \mxsat problems and are known to be highly
symmetric~\cite{kas-ijcai03}.
The approach outlined above was also followed for selecting the instances
to be included in the tables of results.

We have run different publicly available \mxsat solvers, namely {\sc
  MiniMaxSat}~\footnote{http://www.lsi.upc.edu/$\sim$fheras/docs/m.tar.gz},
{\sc
  Toolbar}~\footnote{http://carlit.toulouse.inra.fr/cgi-bin/awki.cgi/ToolBarIntro}
and {\sc
  maxsatz}~\footnote{http://www.laria.u-picardie.fr/$\sim$cli/maxsatz.tar.gz}. ({\sc
  Maxsatz} accepts only plain \mxsat instances.) It has been observed
that {\sc MiniMaxSat} behavior is similar to {\sc Toolbar} and {\sc
  maxsatz}, albeit being in general more robust. For this reason, the
results focus on {\sc MiniMaxSat}.

Tables~\ref{tab:ms} and ~\ref{tab:pms+wms+wpms} provide the results
obtained. Table~\ref{tab:ms} refers to plain \mxsat instances and
Table~\ref{tab:pms+wms+wpms} refers to partial \mxsat (PMS), weighted
\mxsat (WMS) and weighted partial \mxsat (WPMS) instances. For each
instance, the results shown include the number of clauses added as a
result of SBPs (\#ClsSbp), the time required for solving the original
instances (OrigT), i.e. without SBPs, and the time required for
breaking the symmetries plus the time required for solving the
extended formula afterwards (SbpT). In practice, the time required for 
generating SBPs is negligible. The results were obtained on a Intel
Xeon 5160 server (3.0GHz, 1333Mhz, 4MB) running Red Hat Enterprise
Linux WS 4.

\begin{table}[t]
\begin{center}
\caption{Results for {\sc MiniMaxSat} on plain \mxsat instances}
\label{tab:ms}
\begin{tabular}{|l|r|r|r|} 
\hline
Name & \#ClsSbp & OrigT & SbpT \\ \hline \hline
hamming10-2 & 81 & 1000 & 0.19 \\
hamming10-4 & 1 & 886.57 & 496.79 \\
hamming6-4 & 437 & 0.17 & 0.15 \\
hamming8-2 & 85 & 1000 & 0.21 \\
hamming8-4 & 253 & 0.36 & 0.11 \\ \hline
MANN\_a27 & 85 & 1000 & 0.24 \\
MANN\_a45 & 79 & 1000 & 0.20 \\
MANN\_a81 & 79 & 1000 & 0.19 \\ \hline \hline
hole10 & 758 & 42.11 & 0.24 \\
hole11 & 922 & 510.90 & 0.47 \\
hole12 & 1102 & 1000 & 1.78 \\
hole7 & 362 & 0.10 & 0.11 \\
hole8 & 478 & 0.40 & 0.13 \\
hole9 & 610 & 3.68 & 0.17 \\ \hline
Urq3\_5 & 29 & 83.33 & 0.27 \\
Urq4\_5 & 43 & 1000 & 50.88 \\ \hline
chnl10\_11 & 1954 & 1000 & 41.79 \\ 
chnl10\_12 & 2142 & 1000 & 328.12 \\ 
chnl11\_12 & 2370 & 1000 & 420.19 \\ \hline
\end{tabular}
\end{center}
\end{table}

\begin{table}[t]
\begin{center}
\caption{Results for {\sc MiniMaxSat} on partial, weighted and weighted partial \mxsat instances}
\label{tab:pms+wms+wpms}
\begin{tabular}{|l|c|r|r|r|} 
\hline
Name & MStype & \#ClsSbp & OrigT & SbpT \\ \hline \hline
ii32e3 & PMS & 1756 & 94.40 & 37.63 \\
ii32e4 & PMS & 2060 & 175.07 & 129.06 \\ \hline \hline
c-fat500-10 & WMS & 2 & 57.79 & 11.62 \\
c-fat500-1 & WMS & 112 & 0.03 & 0.06 \\
c-fat500-2 & WMS & 12 & 0.16 & 0.11 \\
c-fat500-5 & WMS & 4 & 0.16 & 0.11 \\ \hline
MANN\_a27 & WMS & 1 & 1000 & 880.58 \\
MANN\_a45 & WMS & 1 & 1000 & 530.86 \\
MANN\_a81 & WMS & 1 & 1000 & 649.13 \\ \hline \hline
1502.dir & WPMS & 1560 & 0.34 & 10.67 \\
29.dir & WPMS & 132 & 1000 & 28.09 \\
54.dir & WPMS & 98 & 4.14 & 0.32 \\
8.dir & WPMS & 58 & 0.03 & 0.05 \\ \hline
1502.log & WPMS & 812 & 0.76 & 0.71 \\
29.log & WPMS & 54 & 17.55 & 0.82 \\
404.log & WPMS & 124 & 1000 & 64.24 \\
54.log & WPMS & 48 & 2.37 & 0.16 \\ \hline
\end{tabular}
\end{center}
\end{table}

The experimental results allow establishing the following conclusions:
\begin{itemize}
\item The inclusion of symmetry breaking is {\em essential} for
  solving a number of problem instances. We should note that {\em all}
  the plain \mxsat instances in Table~\ref{tab:ms} for which
  {\sc MiniMaxSat} aborted, are also aborted by {\sc Toolbar} and
  {\sc maxsatz}. After adding SBPs all these instances become 
  easy to solve by any of the solvers.
  For the aborted partial, weighted and weighted partial \mxsat
  instances in Table~\ref{tab:pms+wms+wpms} this is not always 
  the case, since a few instances aborted by {\sc MiniMaxSat} 
  could be solved by {\sc Toolbar} without SBPs. 
  However, the converse is also true, as there are instances that were
  initially aborted by {\sc Toolbar} (although solved by {\sc
    MiniMaxSat}) that are solved by {\sc Toolbar} after adding SBPs.
\item For several instances, breaking only a few symmetries can make
  the difference. We have observed that in some cases the symmetries
  are broken with unit clauses.
\item Adding SBPs is beneficial for most cases where symmetries
  exist. However, for a few examples, SBPs may degrade performance.
\item There is no clear relation between the number of SBPs added and
  the impact on the search time.
\end{itemize}

Overall, the inclusion of SBPs should be considered when a hard
problem instance is known to exhibit symmetries. This does not
necessarily imply that after breaking symmetries the instance becomes
trivial to solve, and there can be cases where the new clauses may
degrade performance. However, in a significant number of cases, highly
symmetric problems become much easier to solve after adding SBPs. In many of
these cases the problem instances become {\em trivial} to solve.


\section{Related Work}
\label{sec:relw}

Symmetries are a well-known research topic, that serve to tackle
complexity in many combinatorial problems.
%
The first ideas on symmetry breaking were developed
in the 90s~\cite{puget-ismis93,crawford-kr96}, by relating symmetries
with the graph automorphism problem, and by proposing the first
approach for generating symmetry breaking predicates. This work was
later extended and optimized for propositional
satisfiability~\cite{kas-ijcai03}. 


Symmetries are an active research topic in
CP~\cite{sym-handbook06}. Approaches for breaking symmetries include
not only adding constraints before search~\cite{puget-ismis93} 
but also reformulation~\cite{smith-leeds01} and
dynamic symmetry breaking methods~\cite{gent-ecai00}.
%
Recent work has also shown the application of symmetries to soft
CSPs~\cite{smith-cp07}.

The approach proposed in this paper for using symmetry
breaking for \mxsat and variants builds on earlier work on symmetry 
breaking for PB constraints~\cite{kas-aspdac04}. 
Similarly to the work for PB constraints, symmetries are identified by
constructing a colored graph, from which graph automorphisms are
obtained, which are then used to generate the symmetry breaking
predicates.


\section{Conclusions}
\label{sec:conc}

This paper shows how symmetry  breaking can be used
in \mxsat and in its most well-known variants, including partial
\mxsat, weighted \mxsat, and weighted partial \mxsat.
Experimental results, obtained on representative instances from the
\mxsat evaluation~\cite{maxsatcomp} and practical
instances~\cite{kas-ijcai03}, demonstrate that symmetry breaking allows
solving problem instances that no state of the art \mxsat solver could 
otherwise solve.
%
For all problem instances considered, the computational effort of
computing symmetries is negligible.
Nevertheless, and as is the case with related work for SAT and PB
constraints, symmetry  breaking should be considered
as an alternative problem solving technique, to be used when standard 
techniques are unable to solve a given problem instance.

The experimental results motivate additional work on symmetry
 breaking for \mxsat. The construction of the
colored graph may be improved by focusing on possible relations among
the different clause weights. Moreover, the use of conditional
symmetries could be considered~\cite{gent-cp05,smith-cp07}.


\subsubsection{\ackname}
This work is partially supported by EU grants IST/033709 and
ICT/217069, by EPSRC grant EP/E012973/1 and by FCT grants
POSC/EIA/61852/2004 and PTDC/EIA/76572/2006.

\bibliographystyle{abbrv}

\end{document}